\documentclass{article}
\usepackage{spconf}

\usepackage[utf8]{inputenc} 
\usepackage[T1]{fontenc}    
\usepackage{url}            
\usepackage{booktabs}       
\usepackage{amsfonts,amsmath,graphicx,nicefrac}
\usepackage[pdfpagelabels]{hyperref}
\usepackage{flushend}

\title{Bias in Automated Image Colorization: Metrics and Error Types}

\name{Frank Stapel \qquad Floris Weers\thanks{The first two authors contributed equally.} \qquad Doina Bucur}
\address{University of Twente}

\begin{document}
\maketitle


\begin{abstract}
We measure the color shifts present in colorized images from the ADE20K dataset, when colorized by the automatic GAN-based DeOldify model. We introduce fine-grained local and regional bias measurements between the original and the colorized images, and observe many colorization effects. We confirm a general desaturation effect, and also provide novel observations: a shift towards the training average, a pervasive blue shift, different color shifts among image categories, and a manual categorization of colorization errors in three classes. 
\end{abstract}



\section{Introduction}

The task of colorizing grayscale images is ambiguous, and difficult for a fully automated process without human input. We provide \emph{systematic measurements of color bias} against the ground truth. We focus on a widely used deep-learning colorizer, DeOldify~\cite{antic2022deoldify} (also a popular Twitter bot), whose colorization models have generative adversarial network (GAN) architectures and are pretrained on ImageNet~\cite{deng2009imagenet}. To measure colorization bias objectively, we take measurements over ADE20K~\cite{zhou2017scene}, a dataset different than the training dataset. 

Colorizers have biases, often demonstrated with example images, not systematic measurements. Early models were accurate on landscape and home, but not on complex scenes~\cite{cheng2015deep,deshpande2015learning}. A frequent problem is \emph{desaturation}~\cite{cheng2015deep,larsson2016learning}, due to loss functions (inherited from standard regression) that encourage conservative predictions. Colorizers aiming at vibrant colors can still fail to recover long-range \emph{color consistency}~\cite{zhang2016colorful,larsson2016learning,su2020instance,antic2022deoldify}, make confusions between colors~\cite{zhang2016colorful}, and output \emph{sepia}~\cite{iizuka2016let,zhang2016colorful,nazeri2018image,ozbulak2019image} or \emph{gray tones}~\cite{isola2017image}. Some models are biased towards frequent colors (e.g., red cars)~\cite{nazeri2018image}, and can confuse the \emph{context or boundaries of objects} (regions with fluctuations may be colored like grassland)~\cite{nazeri2018image,mouzon2019joint,su2020instance,antic2022deoldify}. When an object is not in the GAN distribution (or GAN inversion fails) the output contains unnatural or incoherent colors~\cite{wu2021towards}.

To validate colorization, prior work reports global statistics: pixels-averaged MAE or RMSE~\cite{deshpande2015learning,larsson2016learning,deshpande2017learning,nazeri2018image}), peak signal-to-noise ratio (PSNR)~\cite{cheng2015deep,larsson2016learning,ozbulak2019image,su2020instance,vitoria2020chromagan,wu2021towards}, structural similarity (SSIM)~\cite{ozbulak2019image,su2020instance,wu2021towards}, per-pixel accuracy~\cite{zhang2016colorful,isola2017image,nazeri2018image}, the Fr\'echet inception score (FID)~\cite{wu2021towards}, a colorfulness score~\cite{wu2021towards}, histograms, or human preference. These metrics cannot capture fine-grained biases, such as errors that occur persistently in a certain region of an image category. We thus design not only \emph{global}, but also \emph{local} and \emph{regional bias metrics}. We find a pronounced increase in neutral shades and shift towards the training-average colors, a shift towards blue (pronounced in the center of images), but also that image categories are affected differently (sky patches in nature, urban, and industrial scenes are counterintuitively stripped of blue). In a user study, 60\% of inaccurate colorizations were found to be plausible, with the rest colorization failures.



\section{Method}

DeOldify~\cite{antic2022deoldify} has two colorization models, with different architecture and training process. The \emph{artistic model} creates vibrant, colorful results, but does less well in common scenarios: nature scenes and portraits. The \emph{stable model} is best on nature scenes and portraits, with fewer unnatural miscolorations, but also less vibrancy. Both are trained on a fraction of ImageNet~\cite{deng2009imagenet}. We test here on all (except 5 non-RGB) color images from the ADE20K dataset~\cite{zhou2017scene}. ADE20K provides 25,564 images, split into 10 categories (as in Table~\ref{tab:cat}, column \# images). They are diverse in size and content, are annotated with their semantic category, and the scenes are tagged with objects (such as sky) and object parts.

\begin{table}[htb]
	\caption{{\bf ADE20K image categories} and \% sky}
	\label{tab:cat}
	\vspace{1mm}
	\centering
	{\small
		\begin{tabular}{lrr}
			\textbf{Category}     & \textbf{\# images} & \textbf{\% sky}	\\ \hline
			Urban                 & 7239               & 82.33\%			\\
			Home or hotel         & 6117               &  0.72\%			\\
			Nature landscape      & 3332               & 75.16\%			\\
			Unclassified          & 2536               & 58.49\%			\\
			Workplace             & 1565               &  1.91\%			\\
			Sports and leisure    & 1528               & 41.53\%			\\
			Cultural              & 1115               &  3.23\%			\\
			Shopping and dining   & 1089               &  1.74\%			\\
			Transportation        & 693                &  4.17\%			\\
			Industrial            & 350                & 81.77\%			\\ \hline
		\end{tabular}
	}
\end{table}


To compare original with colorized images, we grayscale (ITU-R 601-2 luma transform~\cite{grayscale_conversion}) and colorize the ADE20K dataset. We then take bias measurements in two color spaces. The {\bf RGB} (trichromatic and additive) color space remains the most widely supported and understood system for the characterization and comparison of colors in digital images. This uses three monochromatic primaries at standardized wavelengths (defined in standard CIE 1931~\cite{CIE152004}), and is perceptually non-uniform: an equal distance in the color space may not correspond to equal differences in color. \textbf{CIELAB} (also L*a*b*), defined in CIE 1976~\cite{CIE152004}, expresses color as three values L* (perceptual lightness), a* (red to green) and b* (blue to yellow). This was intended as a perceptually uniform space: a numerical change in color corresponds to a consistent perceived change in color, so Euclidean distances can be used to compare color transformations in all directions.

\begin{figure}[htb]
	\centering
	\includegraphics[width=\columnwidth]{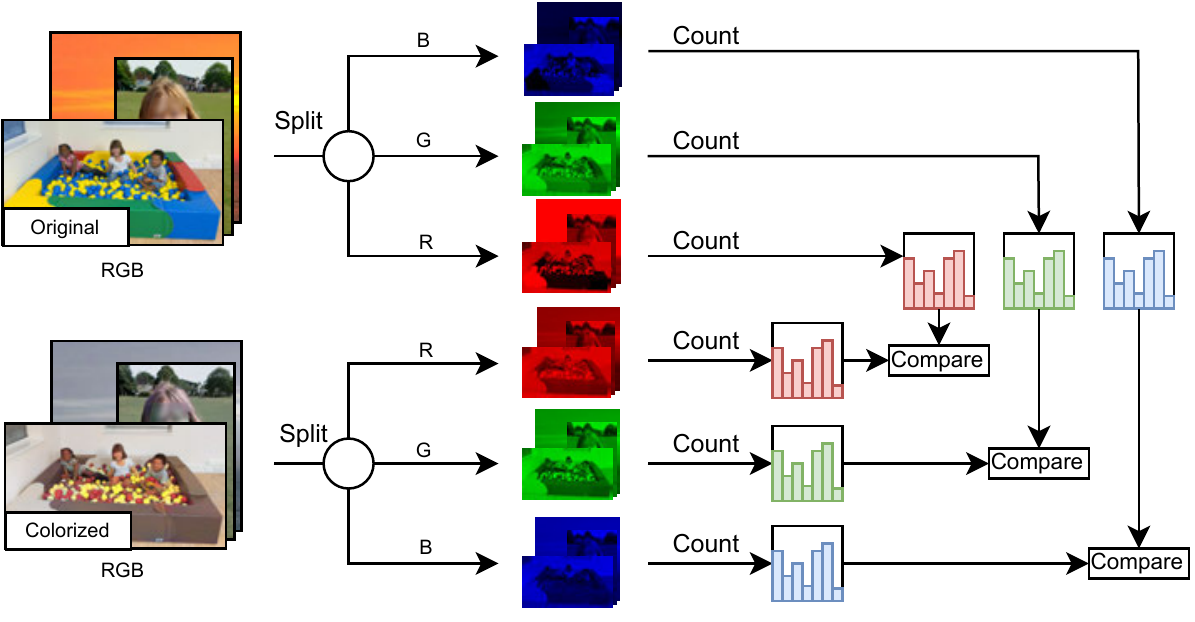}
    \caption{{\bf Global color bias} (method; similar for CIELAB)}
    \label{fig:global_pipeline}
\end{figure}

We systematically take three types of color bias measurements: global, local, and regional, between the original and the colorized images. {\bf Global color bias} measurements show the bias in RGB and CIELAB channel values, treating equally all pixels in all test images. The method for this is shown in Fig.~\ref{fig:global_pipeline}, schematically: the overall distributions of channel values, taken independently per channel, are compared across all images, and we report the channel shifts $\Delta$. This may show, for example, that the pixels in colorized images have high blue-channel values more frequently than the original images. 

\begin{figure}[htb]
	\centering
	\includegraphics[width=\columnwidth]{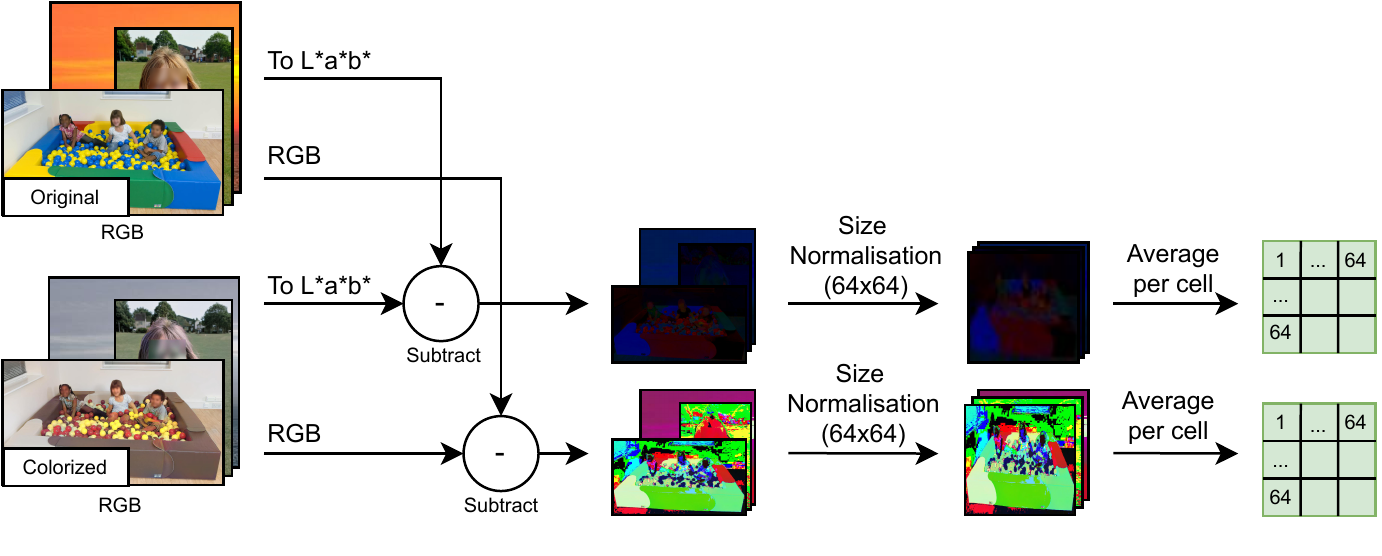}
    \caption{{\bf Local color bias}: color shift (method)}
    \label{fig:local_pipeline}
\end{figure}

{\bf Local color bias} measurements are instead fine-grained. Their aim is to show a two-dimensional color shift between the colorized and the original, on average both for the entire dataset, and per image category. To achieve this, since the ADE20K images are diverse in size, each image, regardless of aspect ratio and resolution, is size-normalized into an aggregated 64x64 image. (This was chosen because it is smaller than the smallest original image size in the dataset.) To avoid confusion between the pixels in the original image and the normalized pixels, we call the latter ``cells''. The color in each cell is the average color of the original pixels. For both RGB and CIELAB, we measure the average color shift per cell. The methodological pipeline for this is shown in Fig.~\ref{fig:local_pipeline}. 

\begin{figure}[htb]
	\centering
	\includegraphics[width=\columnwidth]{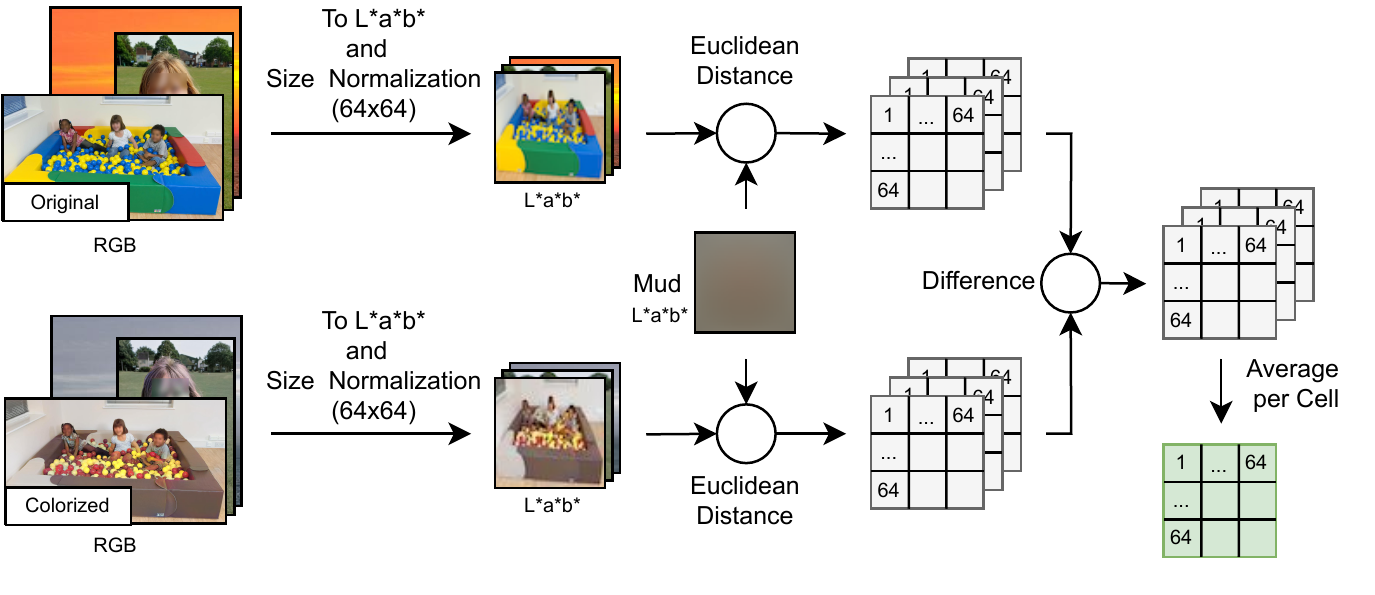}
    \caption{{\bf Local color bias}: distance to mud (method)}
    \label{fig:mud_pipeline}
\end{figure}

We also take a second local bias measurement, which tests the hypothesis whether the colorization strips away some of the vibrant colors and replaces them with dull, muddy shades. For this, we first calculate, per cell in the normalized image size, the average color of the training dataset (2\% of ImageNet). We call this the ``mud'' image. The cells in the mud image need not contain the exact same color. Then, per individual image and per cell, we measure the color distance to the mud color of that cell, and report whether, on average, this distance becomes \emph{smaller} across colorized images. The methodological pipeline is shown in Fig.~\ref{fig:mud_pipeline}. Since this measurement relies on Euclidean distances between colors, we only perform it in the perceptually uniform CIELAB space.

{\bf Regional color bias} measurements are based on the local color shift measurements, but are done separately for specific special regions within the image, such as the center, or the top third. These special regions are image patches defined by popular composition rules in photography~\cite{mavridaki2015comprehensive}, such as the rule to fill the center of the frame with a subject, the \emph{rule of thirds} which defines 3 equal vertical or horizontal patches and their intersection, and the \emph{golden-rule grid}, which does the same in the ratio 1:0.618:1. They are applied as masks to the local color shift results, to draw regional conclusions. 



\section{Results}

We present the measurements, and insights gained from them. All are for DeOldify artistic, unless otherwise specified.

\begin{figure}[htb]
	\centering
    \includegraphics[width=7.2cm]{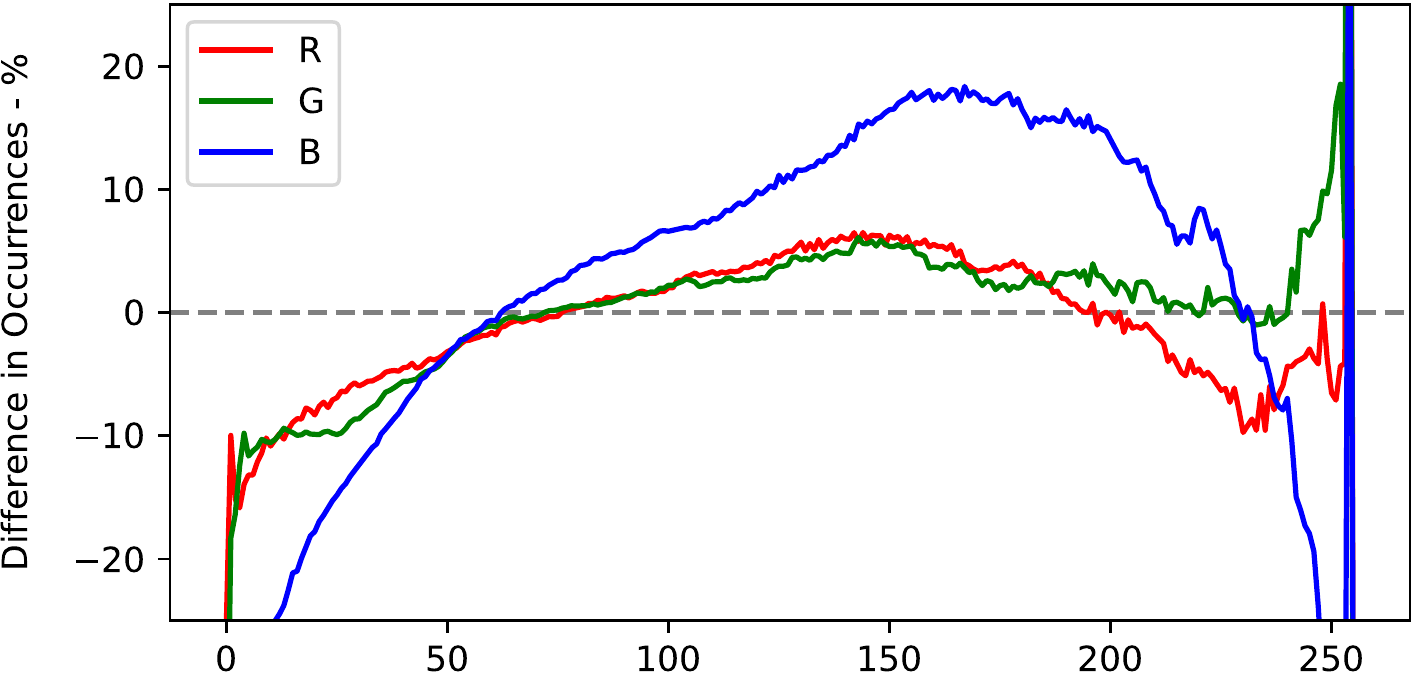}
    \hfill
    \includegraphics[width=7.2cm]{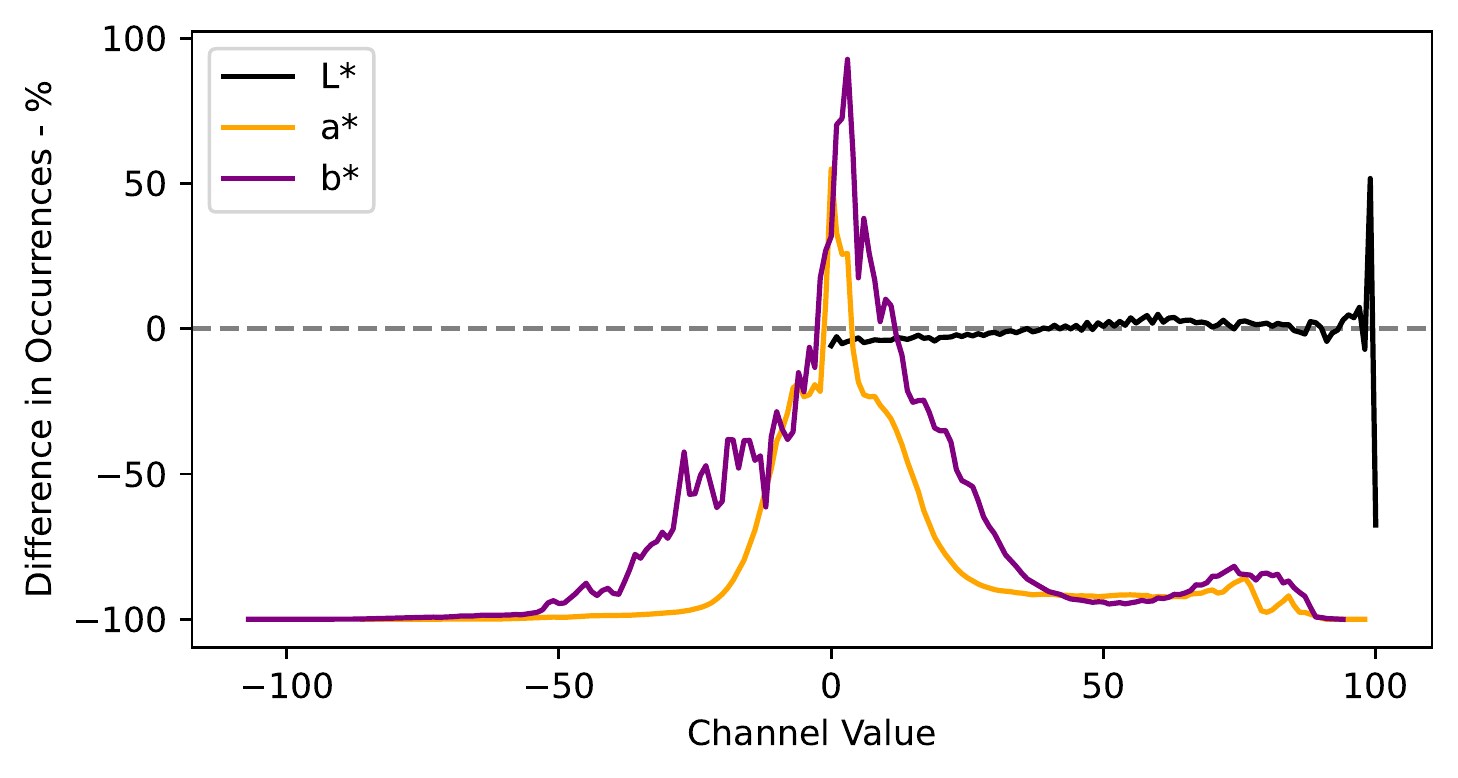}
    \caption{{\bf Global color bias}: channel $\Delta$ (positive means higher frequency in colorized images)}
    \label{fig:global_results}
\end{figure}

{\bf Global color bias.} We show RGB and CIELAB channel shifts $\Delta$ in Fig.~\ref{fig:global_results}. A positive $\Delta$ means a more frequent occurrence of that channel value in the colorized images. (Maximum channel values---R, G, B near 255, L*, a*, b* near 100---are rarely present in the data~\cite{isola2017image}, leading to extreme or noisy shifts at those bounds.) We observe a shift in the distribution of B channel values: an \emph{increased frequency in mid-to-high blue components} among colorized images. This is also present, but is much less significant, for the R, G, and L* channels. On the other hand, the shifts in the distributions of the a* and b* channels show a \emph{pronounced increase in frequency for neutral shades} (between green and red for a*, and blue and yellow for b*), and thus a \emph{pronounced decrease in frequency for saturated colors}. Many a* channel values with an absolute value over 25 disappear from the distribution. In summary, we observe {\bf global bias towards neutral shades, and global bias towards mid-to-high-range blues}.

\begin{figure}[htb]
	\includegraphics[height=2.5cm,width=2.5cm,clip,trim=1mm 1mm 1mm 1mm]{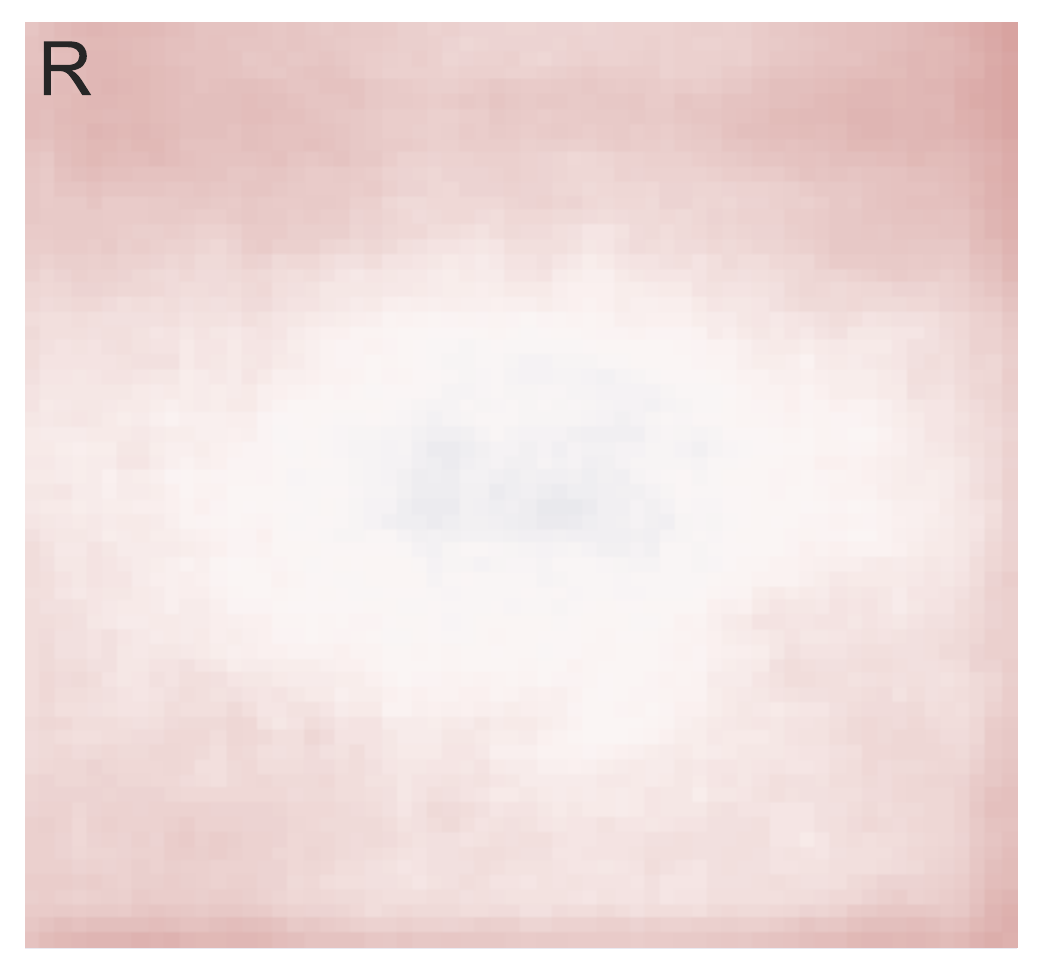}\includegraphics[height=2.5cm,width=2.5cm,clip,trim=1mm 1mm 1mm 1mm]{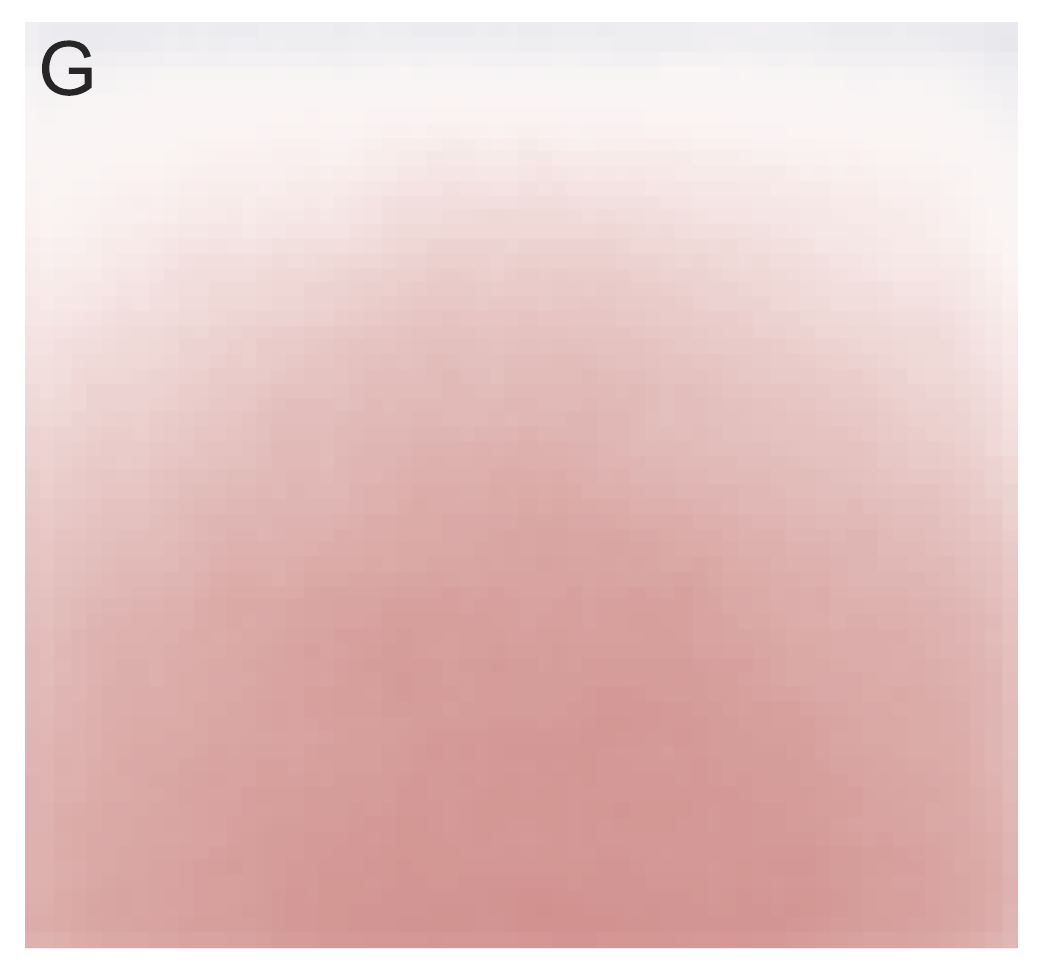}\includegraphics[height=2.5cm,width=3.2cm,clip,trim=1mm 1mm 0mm 1mm]{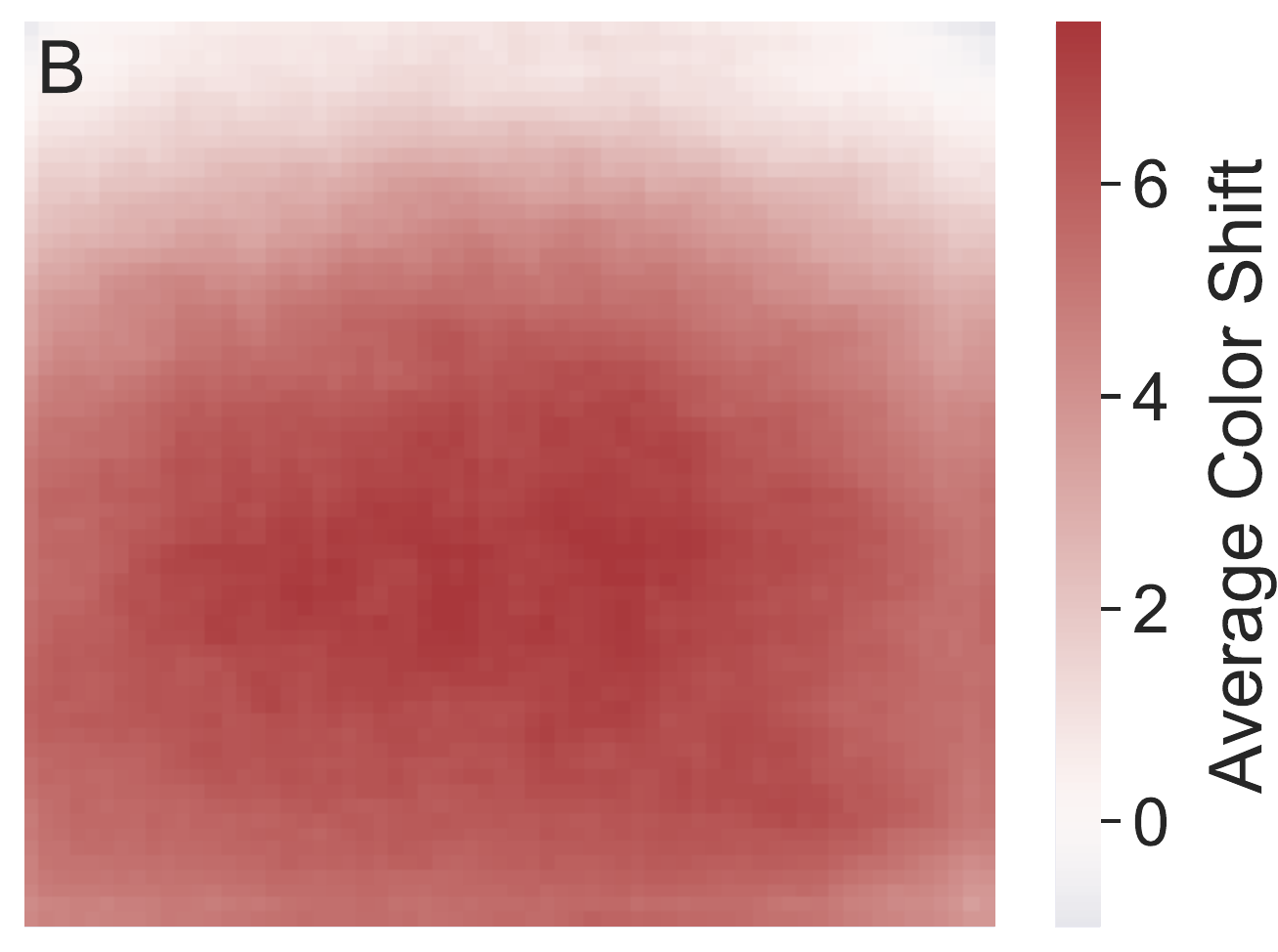}\\\includegraphics[height=2.5cm,width=2.5cm,clip,trim=1mm 1mm 1mm 1mm]{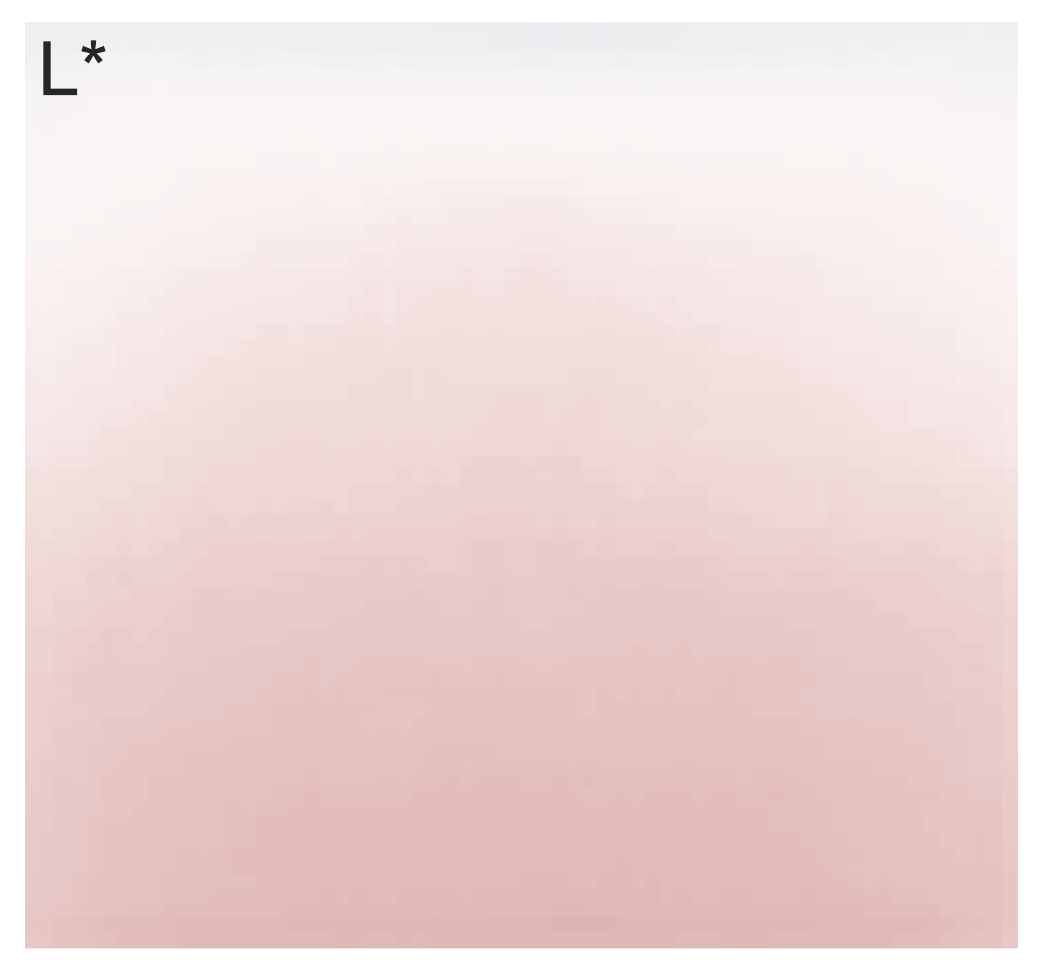}\includegraphics[height=2.5cm,width=2.5cm,clip,trim=1mm 1mm 1mm 1mm]{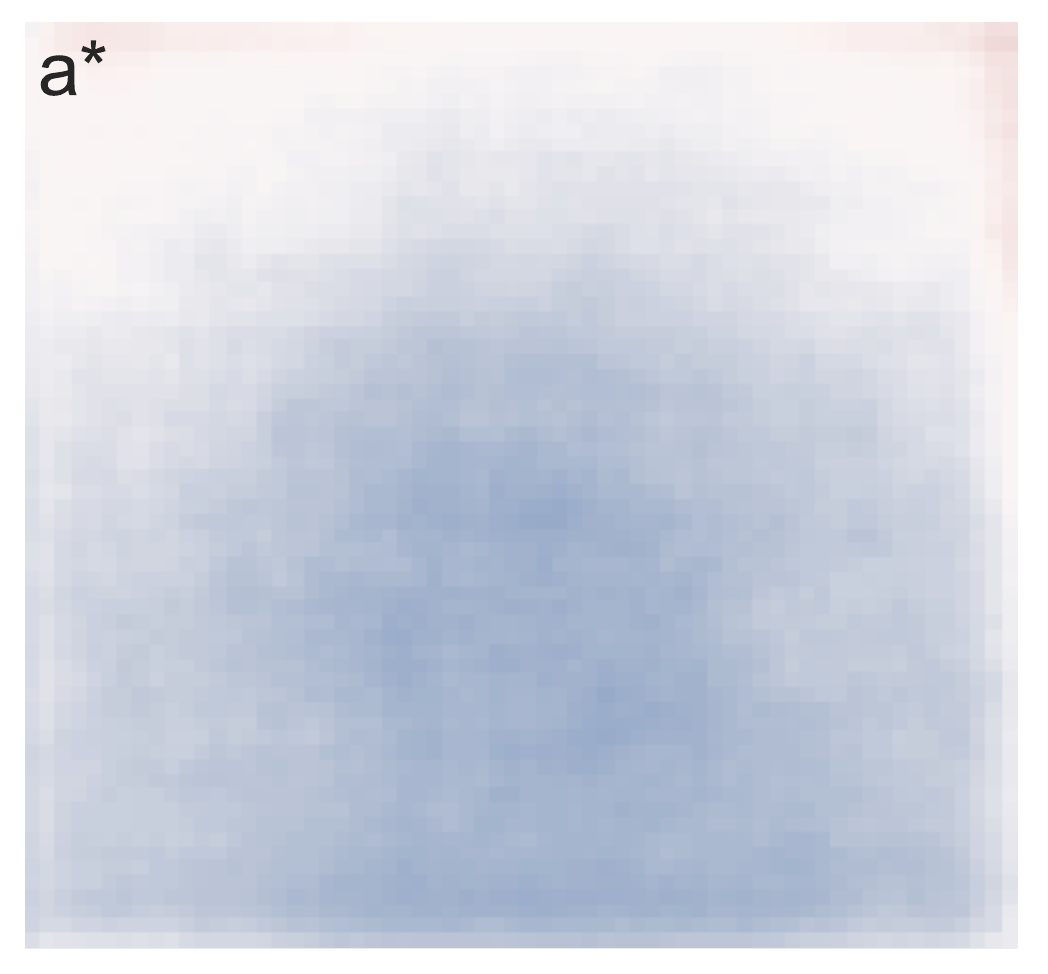}\includegraphics[height=2.5cm,width=3.3cm,clip,trim=1mm 1mm 0mm 1mm]{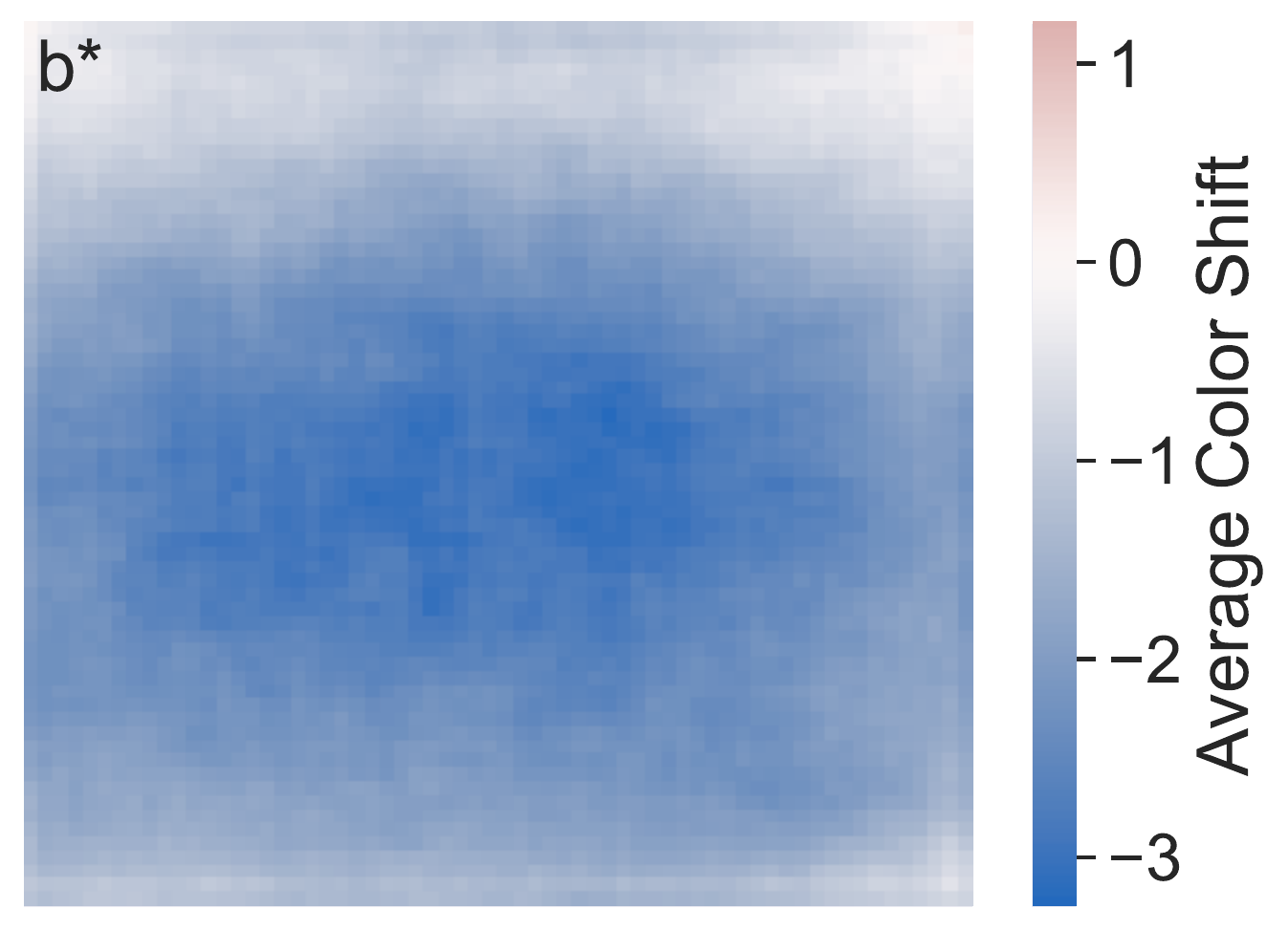}

	\caption{{\bf Local color bias}: average channel shift for R, G, and B (top) and L*, a*, and b* (bottom), over the entire dataset (positive means higher average value in colorized images)} 
	\label{fig:local_results}
\end{figure}

{\bf Local color bias: color shift.} We show per-cell average channel shifts (across the entire dataset) for the R, G, and B  (top) and L*, a*, and b* channels (bottom), in Fig.~\ref{fig:local_results}. A positive per-cell shift means a higher average value in the colorized images. While we had previously (in Fig.~\ref{fig:global_results}) observed similar global shifts between the red and green channels, the local measurements now show that these color shifts have a different spatial distribution: the colorization process \emph{red-shifts slightly the periphery} (but not the center), but \emph{green-shifts slightly the bottom two thirds} of the images. The latter is explained by the more frequent occurrence of natural landscapes at the bottom of the images, with these areas further shifted to green in colorization. The former implies that the (sometimes colorful) objects in the center of the original images are stripped of some of their red, with the opposite occurring for the image background.

\begin{figure}[htb]
	\includegraphics[height=2.5cm,width=2.5cm,clip,trim=2mm 1mm 36.5mm 1mm]{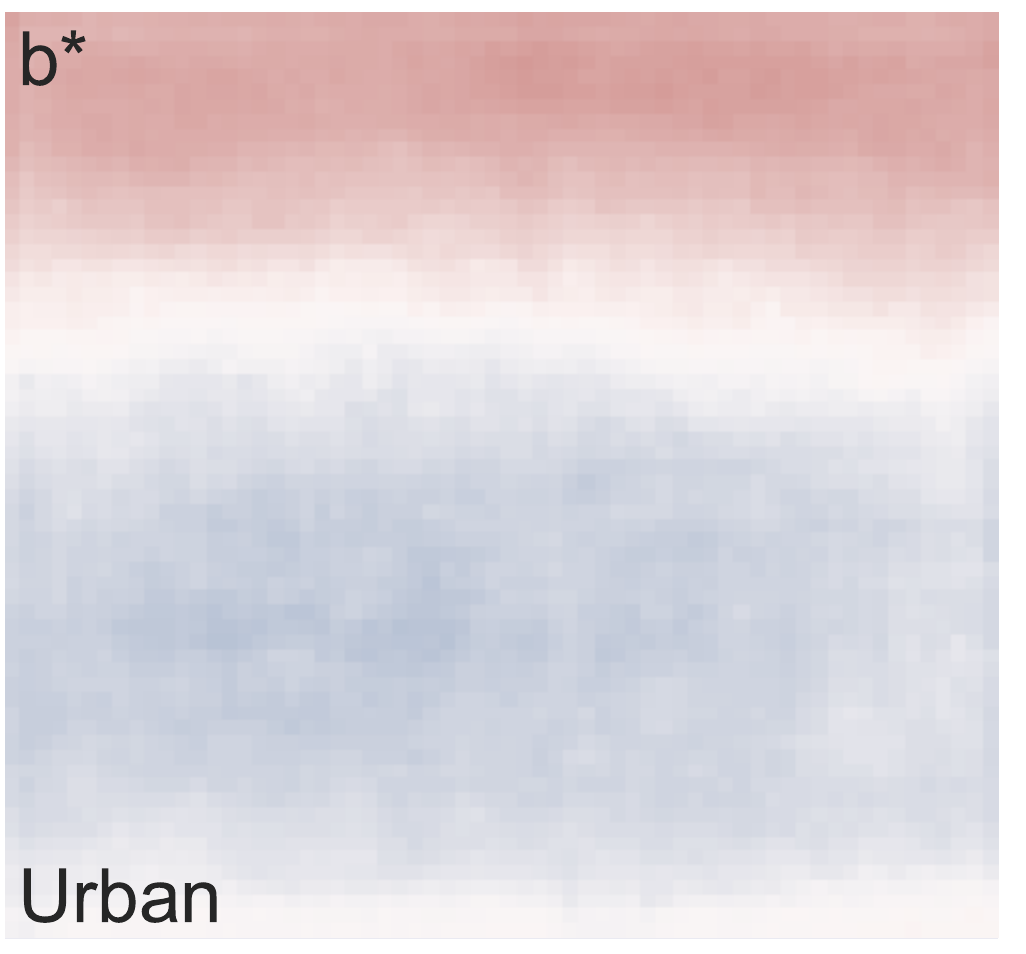}\includegraphics[height=2.5cm,width=2.5cm,clip,trim=2mm 1mm 36.5mm 1mm]{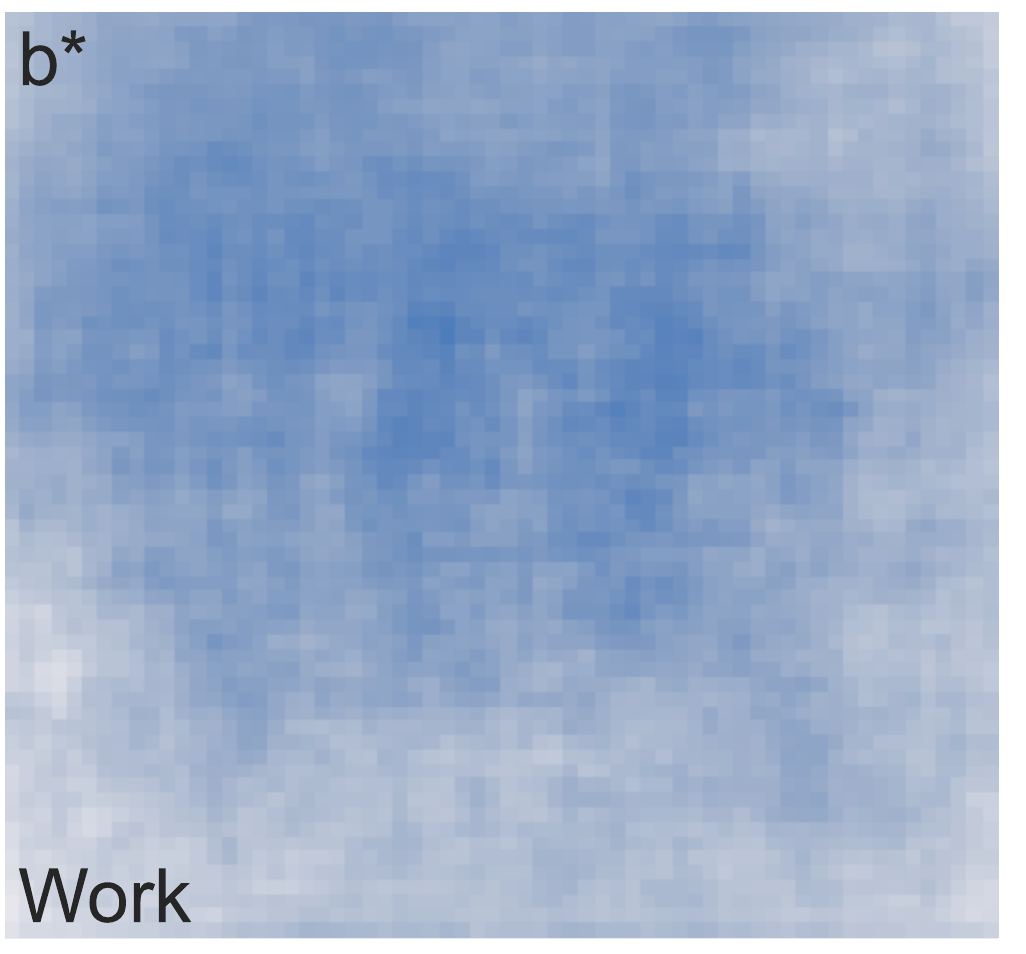}\includegraphics[height=2.5cm,width=3.3cm,clip,trim=2mm 1mm 2.5mm 1mm]{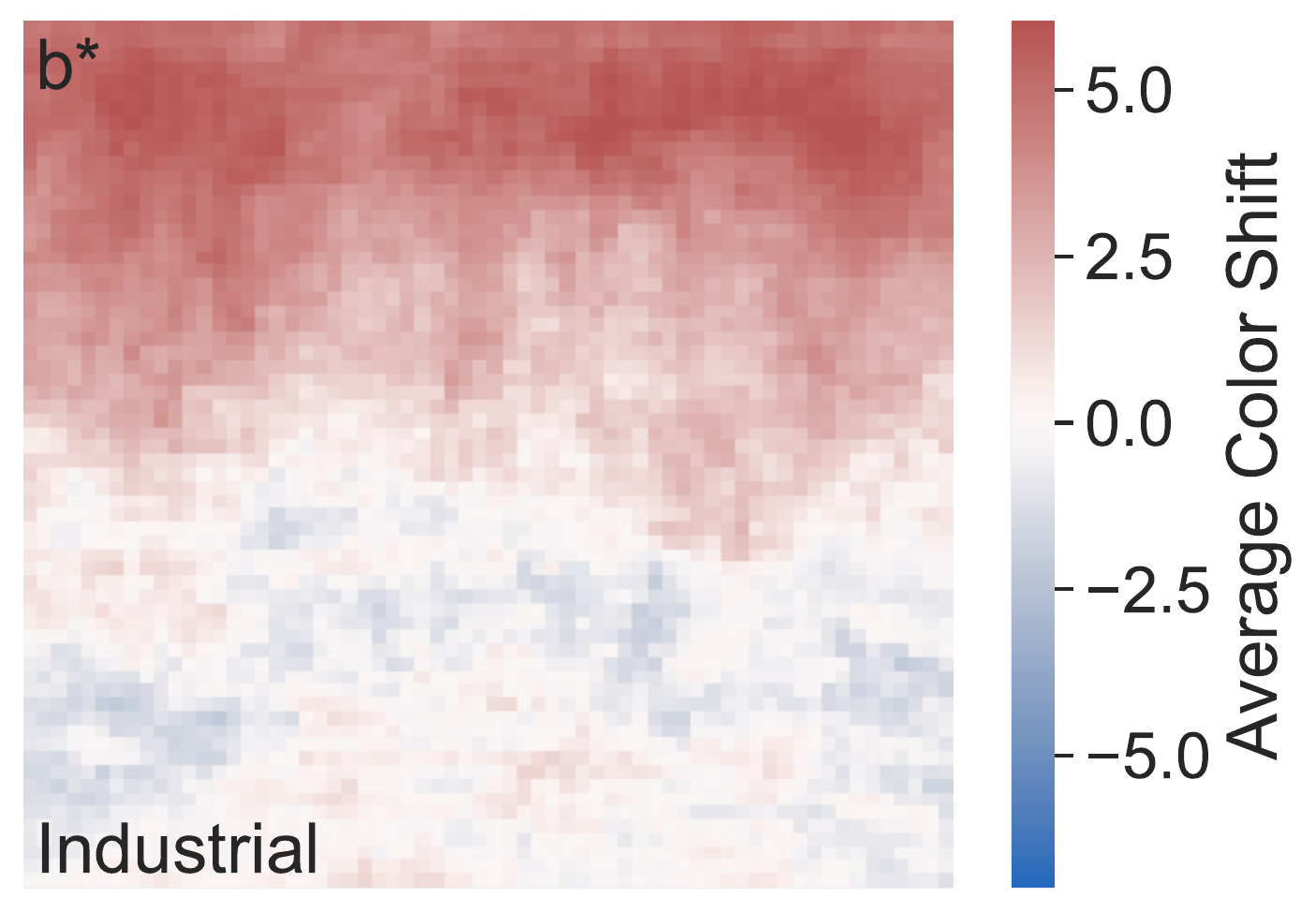}
	\caption{{\bf Local color bias}: b* (blue) channel shift is not uniform per image category (urban, nature, and industrial scenes have top regions shifted away from blue)}
	\label{fig:local_results_blue}
\end{figure}


\begin{figure}[htb]
	\includegraphics[height=2.3cm,width=2.3cm,clip,trim=0mm 0mm 0mm 0mm]{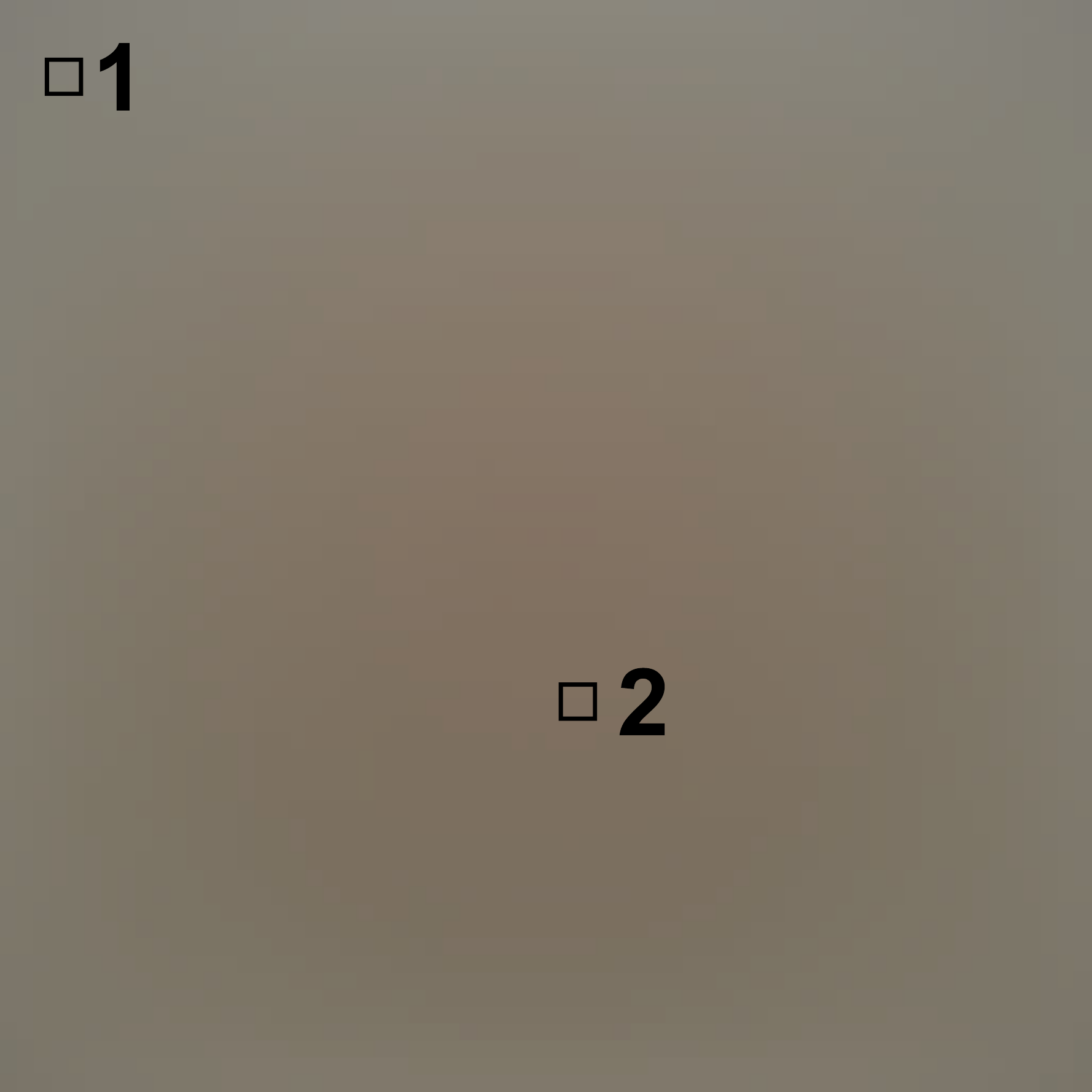}\hfill\includegraphics[height=2.3cm,width=2.9cm,clip,trim=3mm 3mm 2.5mm 3mm]{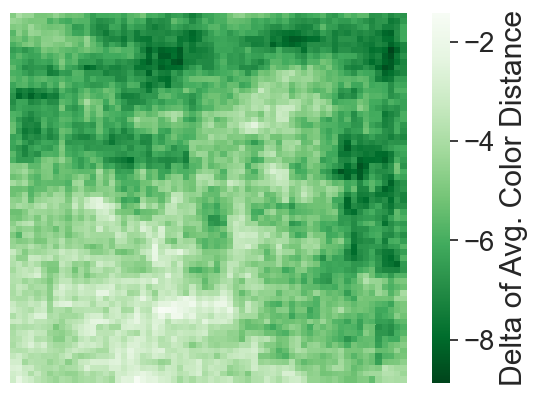}\hfill\includegraphics[height=2.3cm,width=3.3cm,clip,trim=3mm 3mm 2.5mm 3mm]{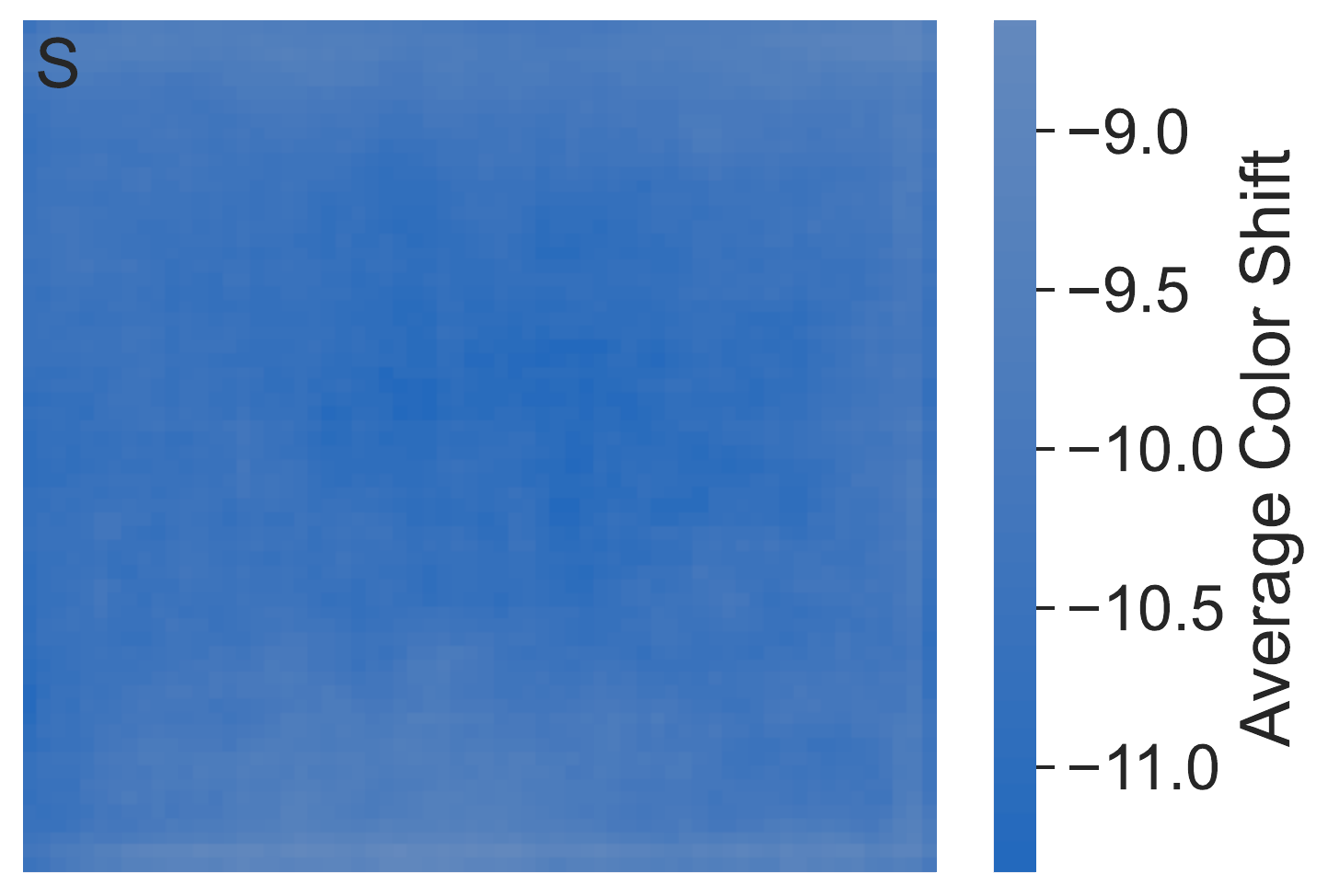}
	\caption{{\bf Local color bias}: (left) mud; (center) difference in distance to mud; (right) average S (saturation) channel shift}
	\label{fig:local_results_mud}
\end{figure}

\begin{figure*}[hbt]
	\centering
	{\footnotesize

	\includegraphics[height=3.2cm]{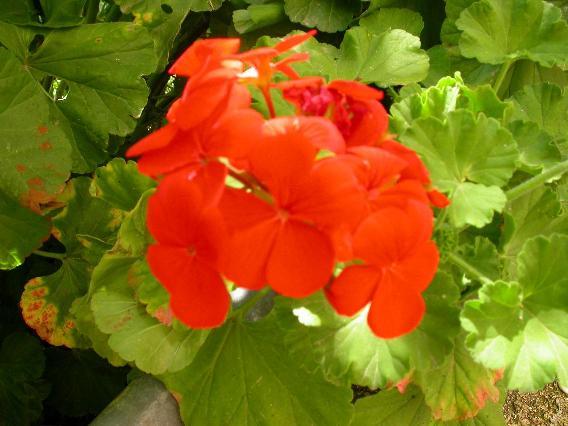}\includegraphics[height=3.2cm]{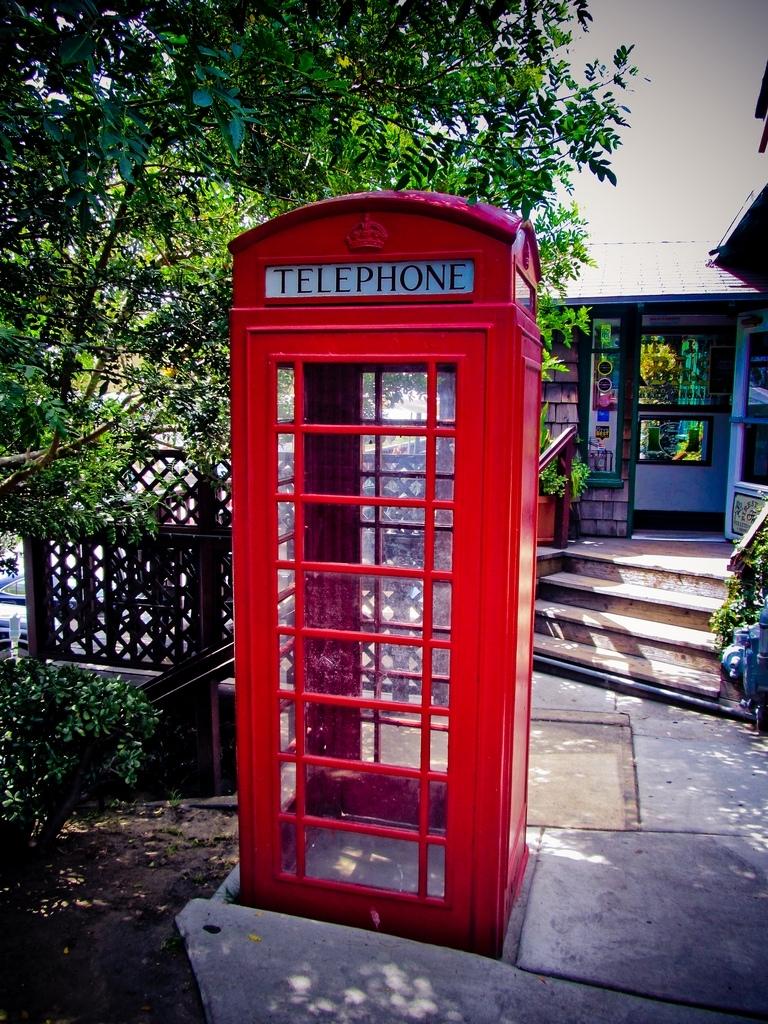}\includegraphics[height=3.2cm]{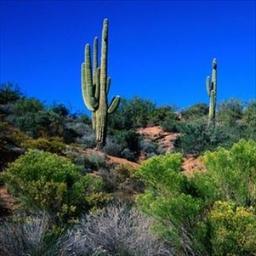}\includegraphics[height=3.2cm]{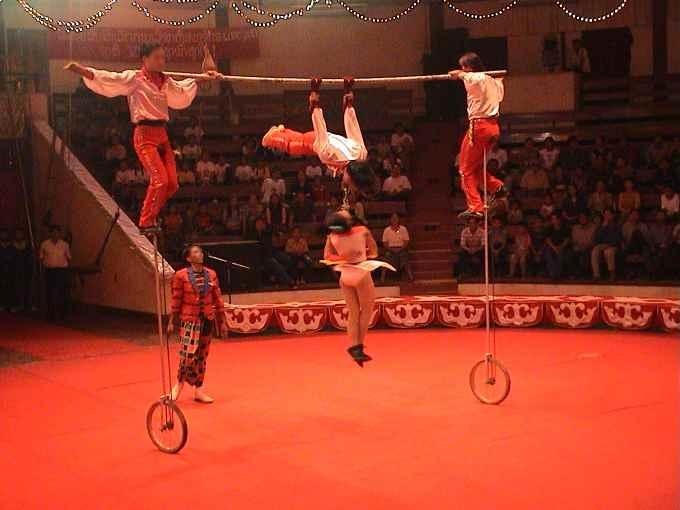}\includegraphics[height=3.2cm]{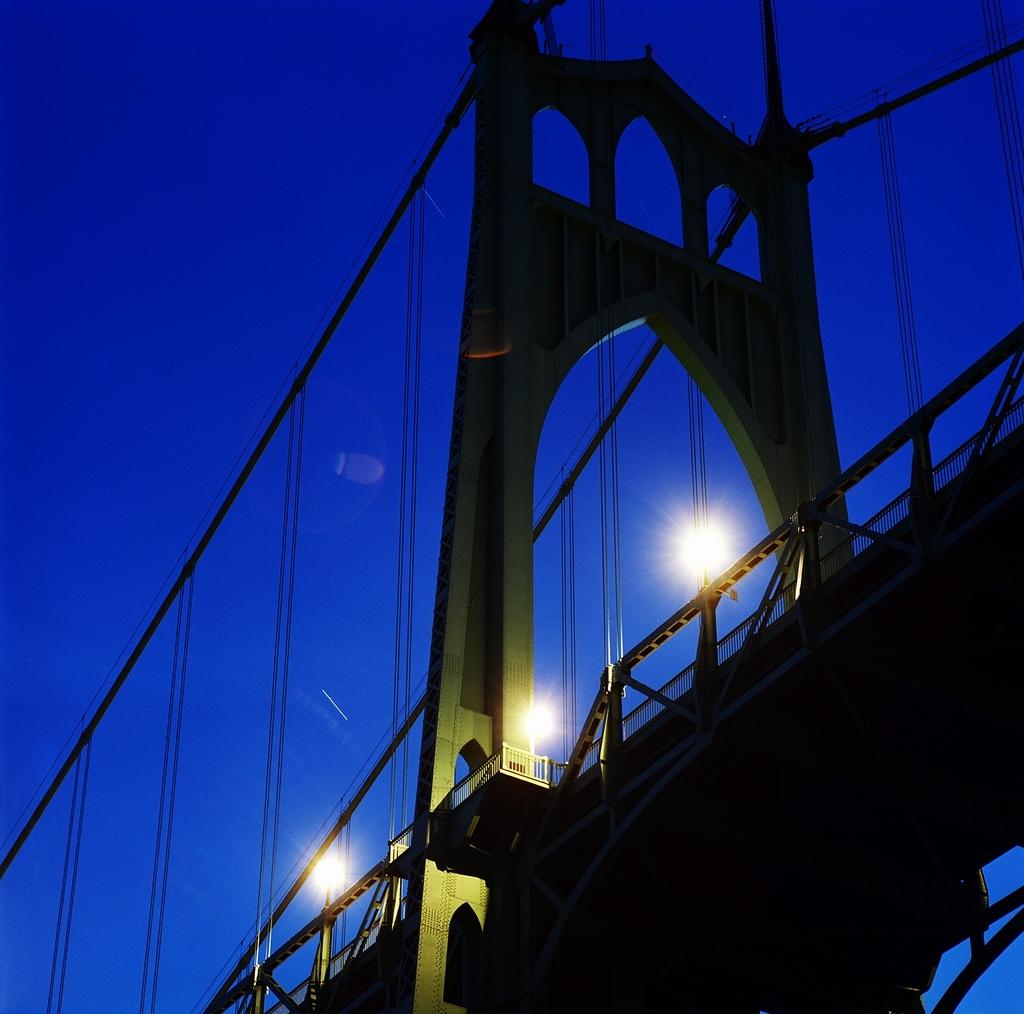}\\[-.5mm]
	\includegraphics[height=3.2cm]{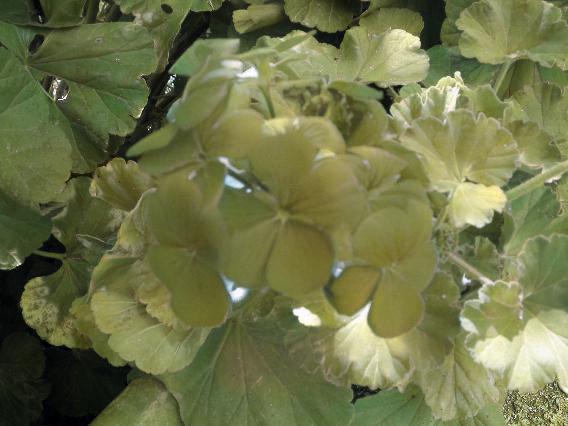}\includegraphics[height=3.2cm]{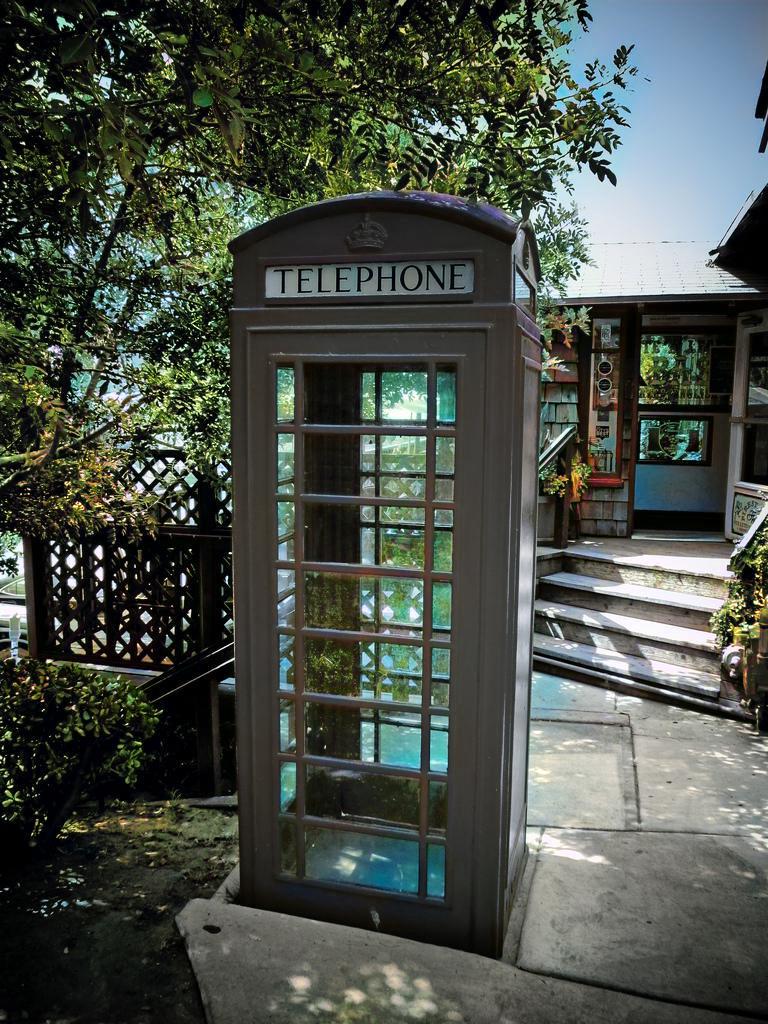}\includegraphics[height=3.2cm]{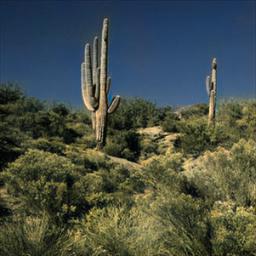}\includegraphics[height=3.2cm]{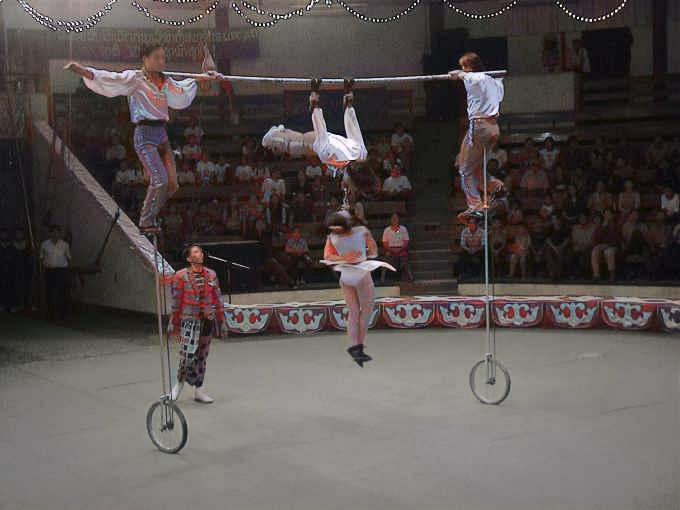}\includegraphics[height=3.2cm]{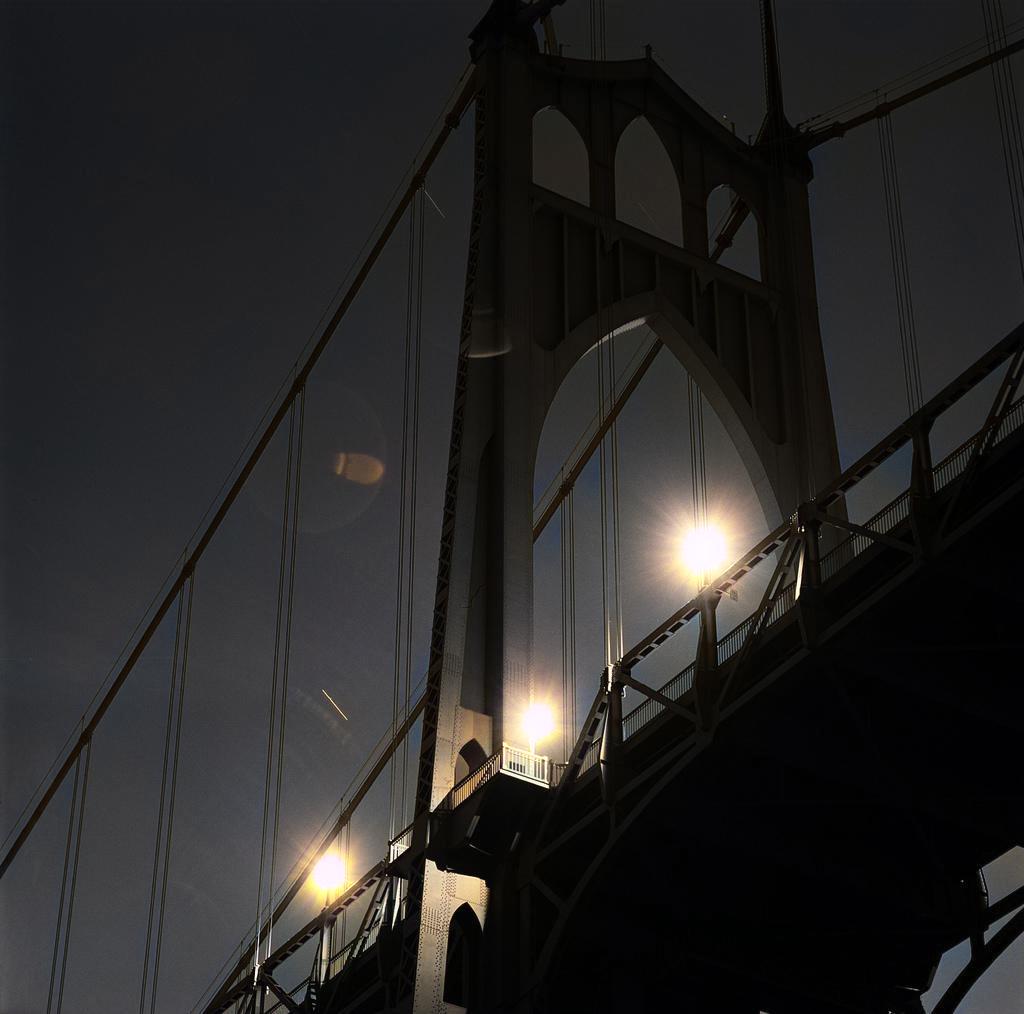}
	
	\begin{minipage}{3.8cm}
	\centering
	(a) Center\\
	(top 1 abs.)
	\end{minipage}
	\begin{minipage}{2.75cm}
	\centering
	(b) Golden ratio (dots)\\
	(top 5 rel.)
	\end{minipage}
	\begin{minipage}{2.95cm}
	\centering
	(c) Rule of Thirds (top)\\
	(top 5 rel.)
	\end{minipage}
	\begin{minipage}{4cm}
	\centering
	(d) Rule of Thirds (bottom)\\
	(top 10 abs.)
	\end{minipage}
	\begin{minipage}{3.2cm}
	\centering
	(e) Rule of Thirds (left)\\
	(top 5 rel.)
	\end{minipage}
	}
    \caption{{\bf Regional color bias}: examples of most inaccurate colorizations (original image at the top)}
    \label{fig:regional}
\end{figure*}

The colorization also on average \textbf{blue-shifts almost every cell} of the images, with double the shift amplitude of the green. The most pronounced shift is in the center of the images. This implies that the blue shift is not a further deepening of the sky blue; it is instead a pervasive effect throughout the images. This is seen in both the B and b* channels: the same blue shift manifests as a positive shift in the B channel, but a negative one (away from yellow and towards blue) in b*. We then verify that this blue shift is \textbf{not uniform across image categories}: Fig.~\ref{fig:local_results_blue} shows the breakdown of the b* channel per image category, for three of the largest categories. (All categories not shown behave like Work, except Nature, which is like Urban.) While 7 out of 10 image categories are almost uniformly blue-shifted, urban, nature, and industrial scenes have their top regions colorized with a shift away from blue. Since these are the image categories with the most sky (present in 75+\% of the images, from Table~\ref{tab:cat}), this means that the colorizer \textbf{strips blue from patches of sky}.

The average Euclidean distance across cells in CIELAB (between colorized and original images) also varies per image category. With lower average distance meaning more accurate colorization, the top are: urban (average distance 1.885), unclassified (2.146), and nature (2.234). The bottom are: work (4.128), shopping (4.143), and cultural (4.331).

{\bf Local color bias: distance to mud.} The mud image (the average color of the training dataset) is shown in Fig.~\ref{fig:local_results_mud} (left). It contains a gradient between cell 1 (RGB 131, 128, 119) and cell 2 (RGB 126, 111, 95). Using this image as a baseline of comparison, we then observe (in Fig.~\ref{fig:local_results_mud}, center) that \emph{the distance image-to-mud decreases} for the artistic model, on average, after colorization: a negative difference in the figure means that the distance is lower on colorized images. The result is very similar for the stable model, but with an even higher amplitude (reaching -10). This confirms our hypothesis and insights in the related work~\cite{cheng2015deep,larsson2016learning,isola2017image} that the colorization strips away some of the colors, and {\bf shifts the colors towards the training average}. This is also confirmed by a supplementary measurement of the average shift in the S channel from the HSV color space (shown in Fig.~\ref{fig:local_results_mud}, right, where the range of channel values is $[0, 100]$). The colorized images are almost \textbf{uniformly and heavily desaturated} on average, in comparison to the originals. 

{\bf Regional color bias.} For each region defined by the rule of thirds and the golden ratio, we extract (1) the top $n$ images by the \emph{absolute} color shift in each region, and (2) the top $n$ images by the \emph{relative} color shift between the region and the rest of the image. The latter allows to capture examples where a region is colorized very differently from the remaining image. Fig.~\ref{fig:regional} shows examples of both. In (a) and (b), the colorization of the dominant object failed, either in absolute or relative terms: (a) the flower was colorized like its leaves, and (b) an object of characteristic color that a human would guess correctly was not colorized. In (c)-(e), the colorization is muted, but plausible. In all, there is overall desaturation.

{\bf Manual categorization of errors.} Finally, we add a user study\footnote{We provide more image examples at \url{https://github.com/WeersProductions/colorization-bias}.}: we selected the top 400 images by (1) the  average and standard deviation in color shifts between original and colorized, and (2) regional color shifts of the relative type. After removing duplicate images in this set, the rest were manually categorised by the type of failure observed in colorization. A minority of the failure modes were not clear-cut. Only 5\% of the images completely failed to colorize and were essentially grayscale. In 23\% of the cases, one dominant object failed to colorize (examples include (a)-(b) in Fig.~\ref{fig:regional}). However, in the majority (60\%) of the cases, the results were judged still plausible (examples include (c)-(e) in Fig.~\ref{fig:regional}), and akin to a ``mood change'' in the image.


\section{Discussion and conclusions}

We presented insights on the color shifts in images colorized by the GAN-based DeOldify model. We introduced local and regional bias measurements between the original and the colorized datasets, and obtained quantitative and qualitative results showing many colorization effects. We observe desaturation (confirming prior knowledge~\cite{cheng2015deep,larsson2016learning,isola2017image}), but also provide \emph{novel observations}: a shift towards the training average, a pervasive blue shift, different shifts among image categories, and a manual classification of the errors. This study has limitations: the measurements included only the two public colorizers available with DeOldify, and only one image dataset---but we conclude that pervasive biases remain present in advanced colorization models. Our results may guide the development of automated AI colorizers, which could, for example, use semantic input to resolve some of the regional color shifts per image category.



\cleardoublepage
\bibliographystyle{IEEEbib}
\bibliography{main}

\end{document}